\documentclass[runningheads]{llncs}
\usepackage{graphicx}
\usepackage{subcaption}
\usepackage{listings}
\usepackage{tabularx,ragged2e}
\usepackage{booktabs}

\begin{document}
\title{The Catalan Language CLUB\thanks{See Acknowledgements for institutional support}}
%
%

\author{Carlos Rodriguez-Penagos,\inst{1}\orcidID{0000-0001-8744-0380} \and
Carme Armentano-Oller\inst{1}\orcidID{0000-0002-9005-4138} \and
Marta Villegas\inst{1}\orcidID{0000-0003-0711-0029}\and
Maite Melero\inst{1}\orcidID{0000-0001-9933-3224}\and
Aitor Gonzalez\inst{1}\orcidID{0000-0002-6106-7810}\and
Ona de Gibert Bonet\inst{1}\orcidID{00000-0002-7163-4807}\and 
Casimiro Carrino Pio\inst{1}\orcidID{0000-0002-2314-9559}} 
\authorrunning{C. Rodriguez-Penagos et al.}
%
\institute{Barcelona Supercomputing Center\\
\email{\{carlos.rodriguez1, carme.armentano, marta.villegas, maite.melero, aitor.gonzalez, ona.degibert, casimiro.carrino,\}@bsc.es}}
\maketitle              
\begin{abstract}
The Catalan Language Understanding Benchmark (CLUB) encompasses various datasets representative of different NLU tasks that enable accurate evaluations of language models, following the General Language Understanding Evaluation (GLUE) example. It is part of AINA and PlanTL, two public funding initiatives to empower the Catalan language in the Artificial Intelligence era.

\end{abstract}
\section{Introduction}
Two public funding initiatives in Spain, PlanTL and AINA, aim at providing Catalan  with the tooling and resources that modern Artificial Intelligence (AI) models can bring to industry, commerce and society in general. These initiatives are enabling high-quality corpora and datasets which, along with extensive Transformer and Spacy\footnote{\url{https://spacy.io/models/ca}} language models, are being made available through various open platforms in order to enable Natural Language Understanding (NLU) capabilities for any institution, organization, company or individual, and not only for big corporations with deep pockets. These efforts have incorporated corpus annotation best practices, and at the same time strive to foster local companies that can deal with the complex tasks needed for Data Science and modern AI. \\
The Catalan Language Understanding Benchmark (CLUB) is a set of test corpora to drive accurate evaluations of language models and downstream applications for real, practical use. As such, they are not modelled after usual linguistic academic research corpus, but are designed for training and evaluating Transformer models, powerful neural network embeddings in widespread use for Artificial Intelligence applications.
\section{Datasets}
We have compiled a new benchmark dataset suite for evaluating NLU capabilities for Catalan, the Catalan Language Understanding Benchmark (CLUB).\footnote{See \url{https://huggingface.co/bsc}} To build it, we bootstrap from existing resources (such as the AnCora  corpus\footnote{\url{http://clic.ub.edu/corpus/en}}) and create new high quality ones from scratch, adopting (and in some cases improving on) existing guidelines. These datasets have permissive licences, and are made  publicly available through the Zenodo platform, under the Catalan AI language resources community,\footnote{\url{https://zenodo.org/}} and through HuggingFace.\footnote{\url{https://huggingface.co/datasets/BSC-TeMU/}}
\subsection{Data statements}
In general, we provide as much curation information as possible  following (Bender and  Friedman, 2018), \cite{tacl_a_00041} when relevant. For example, gender and socioeconomic status are considered not as relevant for the kind of semantic annotations created. However, the fact that all commissioned annotators (1) were native speakers of Catalan, (2) were translators, editors, philologists or linguists, and (3) had previous experience in language-related tasks, \textit{is} considered to be important. The curation rationale we follow was to make these datasets both representative of contemporary Catalan language use, as well as directly comparable to similar reference datasets from the General Language Understanding Evaluation \cite{wang2019glue} (GLUE) benchmark.\footnote{\url{https://gluebenchmark.com/}} Since our datasets are geared towards Machine Learning and Language Modelling, we have provided training, evaluation and tests splits at HuggingFace. In what follows, we describe some of these datasets.
\subsection{POS and NERC from AnCora} 
For \textbf{Part-of-Speech Tagging (POS) }and \textbf{Named Entity Recognition and Classification (NERC)} evaluations, we use annotations from the  of the well-known AnCora corpus, projected on the Universal Dependencies treebank\footnote{\url{https://github.com/UniversalDependencies/UD\_Catalan-AnCora}}. We extracted Named Entities from the original AnCora\footnote{\url{https://doi.org/10.5281/zenodo.4762030}} version, filtering out some unconventional ones, like book titles, and transcribe them into a standard CONLL-IOB format.\\
The AnCora corpus has been released recently under CC-BY licence, and we published our derivative version under the same licence.\footnote{\url{https://huggingface.co/datasets/BSC-TeMU/ancora-ca-ner}} 
\subsection{TECa: Textual Entailment for Catalan} 
Textual entailment (TE) is the directional relation between two sentences, a \textit{premise} and an \textit{hypothesis}. For every \textit{premise} the relation with an \textit{hypothesis} can be Neutral, Inference or Contradiction. Textual Entailment is considered an important evaluation of the ability of a model to incorporate some measure of inferential capabilities, as opposed to make do with a prodigious memory of what it has actually seen before. TECa\footnote{\url{https://doi.org/10.5281/zenodo.4761458}}  contains more than 20\,000 pairs of sentences annotated with their relation label, that can be 0 (neutral), 1 (inference) or 2 (contradiction).\\

\begin{table}
\caption{\label{tab:TE}
Example of a premise and a neutral, inference and contradiction hypothesis.
}
\centering
\begin{tabular}{lp{0.81\linewidth}} 
\hline
Premise: & \textbf{El Port de Barcelona ha guanyat un 30\% el primer semestre del 2016} \\
\hline
Inference: & El Port de Barcelona augmenta la seva activitat \\
Neutral: & El segon semestre de 2016 no ha estat tan bo \\
Contradiction: & L'inici del 2016 ha estat un mal any pel Port de Barcelona \\
\hline
\end{tabular}
\end{table}

Source sentences are extracted from the Catalan Textual Corpus\footnote{\url{https://doi.org/10.5281/zenodo.4519349}}  and from Vilaweb newswire.\footnote{\url{http://www.vilaweb.cat}} \\
We randomly chose 18\,000 sentences from those sources, and filtered them by different criteria, such as length and stand-alone intelligibility. From the remaining text sentences, we commissioned 3 hypotheses (one for each entailment category) to be written by a team of annotators. We obtained more than 20\,000 annotated sentence pairs, which are published under CC-by licence. From 600 randomly selected samples we cross-annotated for Quality Assurance, and obtained an inter-annotator agreement of 83,57\%. 

\subsection{Text Classification: TeCla} 

TeCla\footnote{\url{https://doi.org/10.5281/zenodo.4627197}} (Textual Classification for Catalan) is a Catalan News corpus for thematic Text Classification tasks. It contains 153\,265 articles classified under 30 different categories, albeit editorially-oriented ones rather than truly encyclopedic labels.\\

We crawled 219\,586 articles from the Catalan News Agency (ACN)\footnote{\url{http://www.acn.cat}} newswire archive, the latest from October 11, 2020. We used the \textit{subsection} category as a classification label, after excluding territorial labels and labels with less than 2\,000 occurrences. With this criteria we compiled a total of 153\,265 articles for this text classification dataset.\\

\begin{figure}[!htb]
\centering
    \includegraphics[scale=0.55]{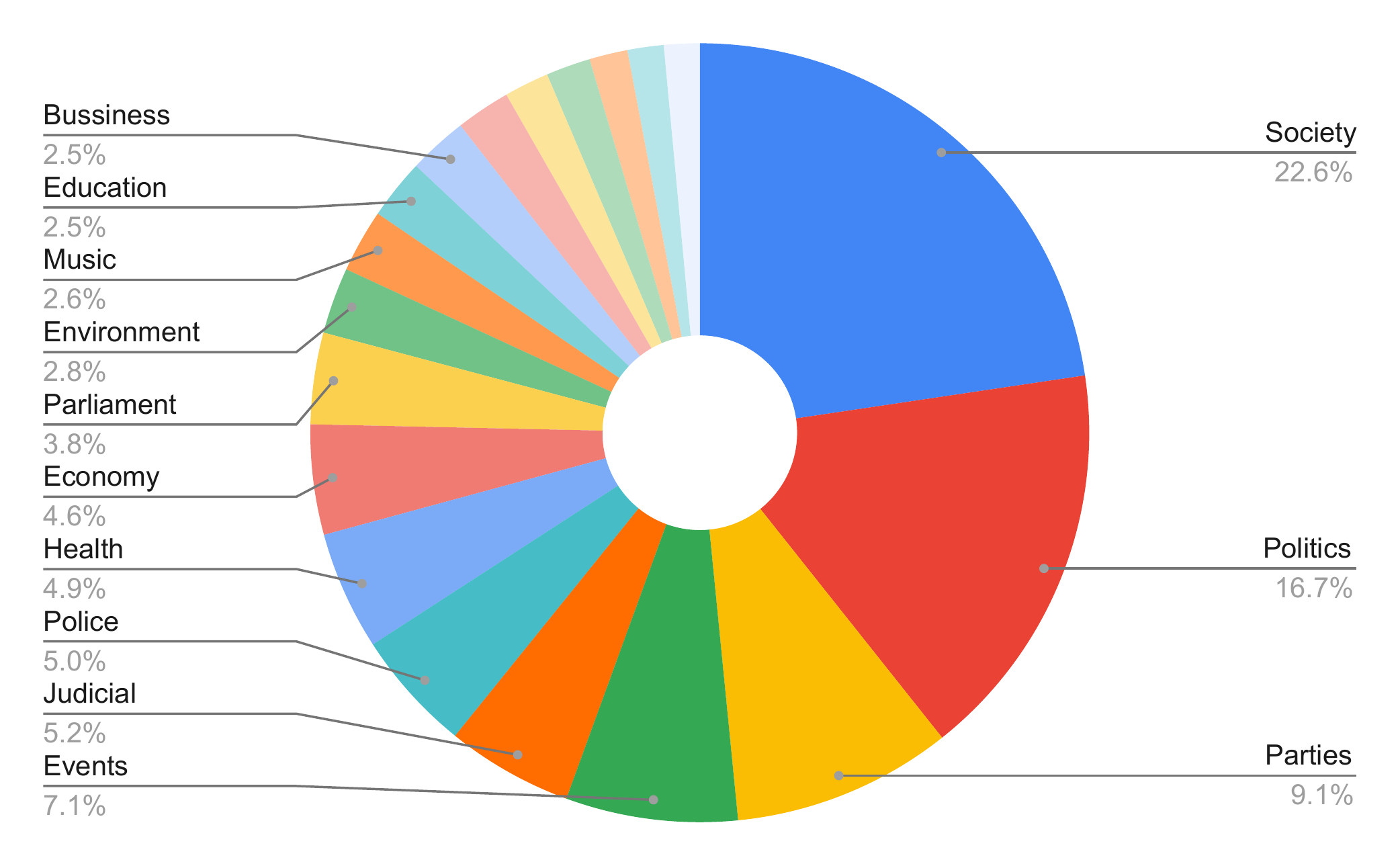}

    \caption{Label distribution of the Text Classification dataset. Here, we show the distribution of the filtered dataset of  labels with at least 2,000 instances.}
    \label{fig:text_classification_dataset}
\end{figure}
\subsection{Extractive Question Answering} 
For Extractive Question Answering we translated or created from scratch three different datasets: XQuAD-Ca, VilaQuAD and ViquiQuad.
\subsubsection{XQuAD-Ca}
The Cross-lingual Question Answering Dataset\footnote{\url{ https://github.com/deepmind/xquad}} is a multilingual benchmark for evaluating  question answering performance. The dataset consists of a subset of 240 paragraphs from the Wikipedia and 1\,190 question-answer pairs from the development set of SQuAD v1.1 (Rajpurkar et al., 2016) together with their professional translations into ten languages: Spanish, German, Greek, Russian, Turkish, Arabic, Vietnamese, Thai, Chinese, and Hindi. Rumanian was added later. We created a Catalan version using professional translators.
\subsubsection{ViquiQuAD}
\label{viquiquad}
Is an extractive Question Answering dataset\footnote{\url{https://doi.org/10.5281/zenodo.4562344}}  using content from the Catalan Wikipedia (Viquipèdia). \footnote{\url{ca.wikipedia.org}} 
From a set of high quality, non-translation, articles in the Catalan Wikipedia, 597 were randomly chosen, and from them  5-8 sentence contexts were extracted. We commissioned the creation of between 1 and 5 questions for each context, following an adaptation of the guidelines from SQUAD 1.0 \cite{rajpurkar2016squad}. Annotators were native language speakers. In total, 15\,153 pairs of a question and an extracted fragment that contained the answer were created.
\subsubsection{VilaQuAD}
Following the same guidelines as in \ref{viquiquad}, we developed VilaQuAD,\footnote{\url{https://doi.org/10.5281/zenodo.4562337}} an extractive QA dataset from newswire. From a the online edition of the catalan newspaper Vilaweb,\footnote{\url{https://www.vilaweb.cat}} 2\,095 article headlines were randomnly selected. These headlines were also used to create the Textual Entailment dataset (TECa). For the extractive QA dataset, creation of between 1 and 5 questions for each news context was commissioned to a team of native speakers. In total, 6\,282 pairs of a question and an extracted fragment that contains the answer were created.

\begin{table}
\centering
\caption{\label{tab:tokens}
Statistics on the number of tokens in contexts, questions, and answers in our QA datasets
}
\begin{tabular}{lrrr} 
\hline
\textbf{ } &  \textbf{XQuAD-ca} & \textbf{ViquiQuAD} & \textbf{VilaQuAD} \\
\hline
Paragraph & 48 & 597 & 2\,095 \\
Context & 240 & 3\,111 & 2\,095\\
Total sentences & 1\,167 & 14\,966 & 11\,462\\
Sentences/context & 4.9 & 4.81 & 5.5\\
Questions & 1\,189 & 15\,153 & 6\,282 \\
Questions/context & 4.9 & 4.9 & 3.0 \\
\hline
Tokens in context & 39\,981 & 468\,250 & 421\,595\\

Tokens in questions & 15\,391 & 145\,124 & 65\,819\\
Tokens in questions/questions & 12.96 & 9.58 & 10.48\\
\hline
Tokens in answers & 4\,424 & 63\,157 & 27\,676\\
Tokens in answers/answers & 3.72 & 4.17 & 4.41\\
  
\hline
\end{tabular}

\end{table}

\begin{table}
\centering
\caption{\label{tab:reasoning}
Question-answer reasoning typology. We sampled 100 random question-answer pairs and classified them manually. We add SQuAD and FQuAD for completeness.
}
\begin{tabular}{lrrrrr} 
\hline
\textbf{ } &  \textbf{XQuAD-ca} & \textbf{ViquiQuAD} & \textbf{VilaQuAD} & \textbf{SQuAD} & \textbf{FQuAD} \\
\hline

   Lexical variation  &   33.0\%    & 7.0\%  & 32.0\%   & 33.3\%   &    35.2\% \\
   World knowledge     &   16.0\%    & 17.0\%   & 16.0\%  &  9.1\% &    11.1\%    \\
   Syntactic variation  &   35.0\%    & 43.0\%  & 22.0\%   &  64.1\%  &   57.4\%  \\
    Multiple sentence   &  17.0\%  & 9.0\%  & 16.0\%    & 13.6\%   &    17.6\%   \\
       
\hline
\end{tabular}
\end{table}

\begin{table*}
\centering
\caption{
Question type frequencies. Differences between XQuAD-ca and XQuAD-en are explained because there are not unambiguous translations for some pronouns. }
\begin{tabular}{llrrrr} 
\hline
\textbf{ } & \textbf{ } &  \textbf{XQuAD-ca} & \textbf{ViquiQuAD} & \textbf{VilaQuAD} & \textbf{XQuAD-en}  \\
\hline

Quin/-s/-a/-es? &  (Which?) & 44.41\%    & 22.42\%  & 20.87 \%  & 7.06\%     \\ 
Què? & (What?) &  16.32\%    & 25.51\%    & 22.59 \% & 57.31\%  \\ 
Qui? & (Who?) &   10.18\%    & 14.75\%    & 18.45 \% & 10.00\%    \\ 
Quant/-s/-a/-es? &  (How many?) & 8.16\%  & 6.1\%   & 9.57 \%  & 6.55\%    \\ 
Com? & (How?) &   7.82\%      & 12.13\%  & 8.21 \% & 5.13\%    \\ 
Quan? & (When?)   &   6.48\%   & 6.5\%     & 7.18 \% & 7.14\%     \\ 
Per què?  &  (Why?) & 1.26\%  & 2.29\%  & 4.36 \%    & 1.26\%      \\ 
On? &  (Where?) & 3.2\%   & 10.16 \% & 8.77 \%   &  3.86\%    \\ 
Other   &    & 2.19\%   & 0.14\%   & 0\%    &   1.93\%  \\
        
\hline
\end{tabular}
\label{tab:question-type}

\end{table*}
\subsection{Semantic Textual Similarity: STS-ca} 

For Semantic Textual Similarity (STS) \cite{agirre-etal-2012-semeval}, we create a new dataset from scratch, STS-ca\footnote{\url{https://doi.org/10.5281/zenodo.4529183}}. It contains more than 3,000 sentence pairs, annotated with their semantic similarity using a scale from 0 (no similarity at all) to 5 (semantic equivalence).\\
To develop STS-ca, we  pre-selected potential sentence pairs from the CaText corpus \footnote{\url{https://doi.org/10.5281/zenodo.4519349}} by using different similarity measures (Jaccard, Doc2Vec and DistilBERT \cite{DBLP:journals/corr/abs-1910-01108} embedding cosine similarity). We did a final manual review to ensure that the selection represented superficial and deep similarities in subject matter and lexicon.  
Following the guidelines set in the SemEval challenges,\footnote{\url{http://ixa2.si.ehu.eus/stswiki}} we commissioned 4 native speaker annotators from 2 independent companies to assess the similarity of the sentence pairs on a scale between 0 (\textit{completely dissimilar}) to 5 (\textit{completely equivalent}), with other possible values, such as 3 (\textit{roughly equivalent, but some important information differs}). Then, for each sentence pair, we computed the mean of the four annotations, and we discarded single annotations that deviate by more than 1 from the mean. After this filtering process, we used the mean of the remaining annotations as a final score. Finally, in order to assess the quality of the dataset, we measured the correlation of each annotator's labels with the average of the rest of the annotators, and averaged all the individual correlations, resulting in a Pearson correlation of 0.739.

\begin{table}
\setlength{\tabcolsep}{5pt}
\caption{Example from STS-ca}\label{tab1}
\begin{tabular}{p{0.45\linewidth}p{0.45\linewidth}c}
\hline
 \textbf{Sentence 1} & \textbf{Sentence 2} & \textbf{STS} \\
\hline
Els manifestants ocupen l'Imperial Tarraco durant una hora fent jocs de taula & Els manifestants ocupen l'Imperial Tarraco i fan jocs de taula & 4 \\
\hline
Aleshores hi posarem un got de vi i continuarem amb la cocció fins que s'hagi evaporat el vi i ho salpebrarem. & Mentre, hi posarem el vi al sofregit i deixarem coure uns 7/8, fins que el vi s'evapori. & 3 \\
\hline
\end{tabular}
\end{table}

\section{Evaluating Transformer models using CLUB}
CLUB was created originally to evaluate the biggest embeddings yet deployed for the Catalan language, using huge crawlings of the .cat domains, along with specific ones of catalan government websites, resulting in a RoBERTa-base model  (BERTa)\footnote{\url{https://huggingface.co/PlanTL-GOB-ES/roberta-base-ca}} introduced in \cite{armengolestape2021multilingual}. This model was created with more than 1\,700 million word tokens, and outperforms in these tasks all other mono or multilingual models available for this Iberian language. 

For our model, we leverage the Huggingface Transformers library \cite{wolf2020huggingfaces}. For each task, we attach a linear layer to the models and fine tuned with the training set of the specific dataset. We trained each specific model under the same settings across tasks consisting of 10 training epochs, with an effective batch size of 32 instances, a learning rate of $5e^{-5}$ and a maximum input length of 512 tokens for all the datasets but TECA, where we used 128 tokens. The rest of the hyperparameters are set to the default values in Huggingface Transformers. We select the best checkpoint as per the task-specific metric in the corresponding validation set, and then evaluate it on the test set. We report the results and metrics used in Table \ref{tab:results-tasks}.\footnote{We use F1 for POS and NERC, accuracy for TC, an average of Pearson and Spearman (Pe/Sp) coefficient for STS and F1/Exact Match for QA (ViquiQuAD).}  \\

{\tabcolsep=0pt\def\arraystretch{1.1}
\captionof{table}{Results of BERTa Catalan model using CLUB. }\label{tab:results-tasks}
\begin{tabularx}{350pt}{c *6{>{\Centering}X}}\toprule
  \multicolumn{2}{c}{\textbf{QA viquiquad}} &  \textbf{STS} & \textbf{NERC} & \textbf{POS} & \textbf{TE}
  \tabularnewline \cmidrule(lr){1-2}
  F1 & EM & Pe/Sp & F1 & F1 & Acc \tabularnewline \midrule
  86.99 & 73.24& 81.56 & 89.63 & 98.94 & 79.12 
  \tabularnewline \bottomrule
\end{tabularx}}

\section{Further work}
Project AINA will ensure that Catalan has the resources and models needed to claim its rightful place in a global digital AI ecosystem. We are also designing and commissioning other datasets, such as realistic chatbot conversations for customer support, Winograd tests, Sentiment Analysis corpora, etc. Also forthcoming are Speech corpus, like the recently published ParlamentParla \footnote{\url{https://zenodo.org/record/5541827}} (more than 600 hours of speech from Catalan Parliament sessions) and contributions to the Catalan Commonvoice campaigns.\footnote{\url{https://commonvoice.mozilla.org}}. You can follow AINA through its official website. \footnote{\url{https://politiquesdigitals.gencat.cat/en/tic/aina-el-projecte-per-garantir-el-catala-en-lera-digital/}}

\section{Acknowledgments}
This work was partially funded by the Generalitat de Catalunya through the project PDAD14/20/00001, the State Secretariat for Digitalization and Artificial Intelligence (SEDIA) within the framework of the Plan TL,\footnote{\url{https://www.plantl.gob.es/}} and the MT4All CEF project.

%
%
%
\bibliographystyle{splncs04}
%
\bibliography{OpenCor2021}

\end{document}